\documentclass{article}
\usepackage{spconf,amsmath,graphicx}

\usepackage{graphicx}
\usepackage{amsmath}
\usepackage{amsthm}
\usepackage{booktabs}
\usepackage{algorithm} 
\usepackage{algorithmic}
\usepackage{multirow}
\usepackage{subfigure}

\title{Training Robust Spiking Neural Networks with 
ViewPoint Transform and SpatioTemporal Stretching}
\name{Haibo Shen\textsuperscript{1}, 
Juyu Xiao\textsuperscript{1}, 
Yihao Luo\textsuperscript{2,1}, 
Xiang Cao\textsuperscript{3,1}, 
Liangqi Zhang\textsuperscript{1},  
Tianjiang Wang\textsuperscript{1}\thanks{This work was supported in part by the National Natural Science Foundation of China under Grant 61572214 and Seed Foundation of Huazhong University of Science and
Technology (2020kfyXGYJ114). (Corresponding author: Tianjiang Wang.)}}
\address{School of Huazhong University of Science and Technology\textsuperscript{1}\\
Yichang Testing Technique Research Institute\textsuperscript{2}\\
Changsha University\textsuperscript{3}}
\begin{document}
%\ninept
%
\maketitle
\begin{abstract}
Neuromorphic vision sensors (event cameras) 
simulate biological visual perception systems and 
have the advantages of high temporal resolution, 
less data redundancy, low power consumption, and 
large dynamic range. 
Since both events and spikes are modeled from neural signals, 
event cameras are inherently suitable
for spiking neural networks (SNNs), which are considered 
promising models for artificial intelligence (AI) and 
theoretical neuroscience. However, the unconventional 
visual signals of these cameras pose a great challenge 
to the robustness of spiking neural networks. 
In this paper, we propose a novel data augmentation method, 
ViewPoint Transform and SpatioTemporal Stretching (VPT-STS). 
It improves the robustness of SNNs by transforming 
the rotation centers and angles in the spatiotemporal domain 
to generate samples from different viewpoints. 
Furthermore, we introduce the spatiotemporal stretching 
to avoid potential information loss in viewpoint transformation. 
Extensive experiments on prevailing neuromorphic datasets demonstrate
that VPT-STS is broadly effective 
on multi-event representations and significantly outperforms 
pure spatial geometric transformations. 
Notably, the SNNs model with VPT-STS achieves a 
state-of-the-art accuracy of 84.4\% on the DVS-CIFAR10 dataset.
\end{abstract}

\section{Introduction}
Inspired by the primate visual system, 
neuromorphic vision cameras generate events by 
sampling the brightness of objects. For example, 
the Dynamic Vision Sensor (DVS)~\cite{DVS} camera 
and the Vidar~\cite{Vidar} camera
are inspired by the outer three-layer structure of the retina
and the foveal three-layer structure, respectively.
Both of them have the advantages of high temporal resolution, 
less data redundancy, low power consumption, 
and large dynamic range~\cite{EventSurvey}. 
In addition, spiking neural networks (SNNs) are similarly inspired 
by the learning mechanisms of the mammalian brain and are considered 
a promising model for artificial intelligence (AI) and theoretical 
neuroscience~\cite{SNN}. In theory, as the third generation of 
neural networks, SNNs are computationally more powerful than 
traditional convolutional neural networks (CNNs)~\cite{SNN}.
Therefore, event cameras are inherently suitable for 
SNNs.

However, the unconventional visual signals of these 
cameras also pose a great challenge to the 
robustness of SNNs.
Most existing data augmentations are fundamentally 
designed for RGB data and lack exploration of neuromorphic 
events. For example,  
% translation~\cite{ShortenK19Translation} moves the image around 
% to avoid positional biases in the data.
Cutout~\cite{TerranceCutOut} artificially impedes a 
rectangular block in the image to simulate 
the impact of occlusion on the image. 
Random erasing~\cite{randomerasing} further 
optimizes the erased pixel value by adding noise. 
Mixup~\cite{ZhangCDL18Mixup} uses the weighted 
sum of two images as training samples to smooth 
the transition line between classes.
Since neuromorphic data have an additional temporal 
dimension and differ widely in imaging principles, 
novel data augmentations are required to process 
the spatiotemporal visual signals of these cameras.

In this paper, we propose a novel data augmentation method 
suitable for events,
ViewPoint Transformation and SpatioTemporal Stretching (VPT-STS).
Viewpoint transformation solves the spatiotemporal scale mismatch 
of samples by introducing a balance coefficient, 
and generates samples from different viewpoints 
by transforming the rotation centers and 
angles in the spatiotemporal domain. 
Furthermore, we introduce spatiotemporal stretching 
to avoid potential information loss in viewpoint transformation. 
Extensive experiments are performed on prevailing neuromorphic datasets. 
It turns out that VPT-STS 
is broadly effective on multiple event representations
and significantly outperforms pure 
spatial geometric transformations. 
Insightful analysis shows that VPT-STS improves the robustness 
of SNNs against different spatial locations. 
In particular, the SNNs model with VPT-STS achieves 
a state-of-the-art accuracy of 84.4\% on the DVS-CIFAR10 dataset.
 
Furthermore, while this work is related to EventDrop~\cite{GuSHY21}, 
NDA~\cite{NDA}, there are some notable differences. 
For example, NDA is a pure global geometric transformation, 
while VPT-STS changes the viewpoint of samples in the spatiotemporal 
domain. EventDrop is only experimented on CNNs, 
it introduces noise by dropping events, 
but may cause problems with dead neurons on SNNs. 
VPT-STS is applicable to both CNNs and SNNs, 
maintaining the continuity of samples. 
In addition, EventDrop transforms both temporal 
and spatial domains, but as two independent strategies, 
it does not combine the spatiotemporal information of the samples. 
To our knowledge, VPT-STS is the first event data augmentation 
that simultaneously incorporates spatiotemporal transformations.

\section{Method}\label{sec:method}

\subsection{Event Generation Model}
The event generation model~\cite{EventSurvey,SNN} 
is abstracted from dynamic vision sensors~\cite{DVS}. 
Each pixel of the event camera responds to changes in 
its logarithmic photocurrent $L=\log(I)$. Specifically, 
in a noise-free scenario, an event $e_k = (x_k, y_k, t_k, p_k)$ is 
triggered at pixel $ X_k = (y_k, x_k)$ and at time $t_k$ as soon as the brightness 
variation $|\Delta L|$ reaches a temporal contrast threshold $C$ 
since the last event at the pixel.
The event generation model can be expressed by the following formula:
\begin{equation}\label{event_generate}
    \begin{aligned}
        \Delta L(X_k, t_k) = L(X_k, t_k) - L(X_k, t_k - \Delta t_k) = p_k C
    \end{aligned}
\end{equation}
where $C > 0$, $\Delta t_k$ is the time elapsed since the last event at
the same pixel, and the polarity $p_{k} \in\{+1,-1\}$ is the sign
of the brightness change. During a period, 
the event camera triggers event stream $\mathcal{E}$:
\begin{equation}\label{events_stream}
    \mathcal{E}=\left\{e_{k} \right\}_{k=1}^{N}=\left\{\left(X_{k}, t_{k}, p_{k}\right)\right\}_{k=1}^{N}
\end{equation}
where $N$ represents the number of events in the set $\mathcal{E}$. 

\begin{figure}[tb]
  \centering
    \centerline{\includegraphics[width=7.0cm]{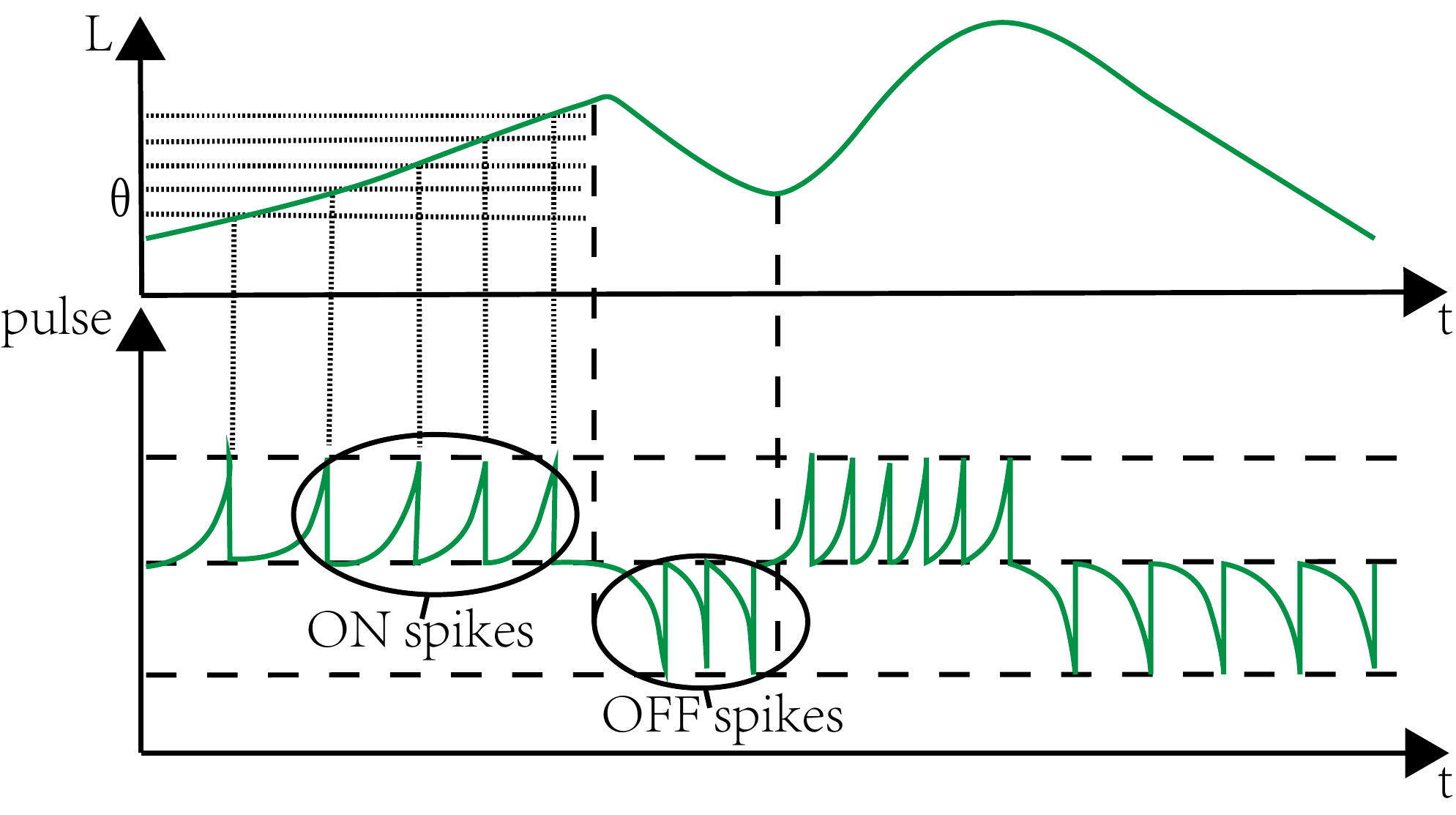}}
  \caption{Event generation model.}
  \label{fig_Sampling}
\end{figure}

As shown in Figure~\ref{fig_Sampling},
an event is generated each time the brightness variances reach the threshold, 
and then $|\Delta L|$ is cleared. 
The event stream can be represented as a matrix:
\begin{equation}\label{eq_event_stream_matrix}
  \begin{aligned}
    M_{\varepsilon} = 
    \begin{pmatrix}
      y_1 & x_1 & t_1 & 1 \\
      \vdots & \vdots & \vdots & \vdots \\
      y_N & x_N & t_N & 1 \\
    \end{pmatrix}_{4 \times N}
  \end{aligned}
\end{equation}

For convenience, we omit the unconverted polarity $p$.

\subsection{Motivation}
This work stems from the observation that it is 
difficult to maintain absolute frontal view between 
the sample and cameras, 
which easily leads to a slight shift of the viewpoint. 
Considering this small offset distance, 
we use viewpoint rotation to approximate the deformation 
of samples in space and time. 
In addition, since events record the brightness change 
of samples, especially changes of the edge, 
variations of the illumination angle will also cause 
the effect of viewpoint transformation, which suggests 
that we can enhance the robustness of SNNs by 
generating viewpoint-transformed samples.

\subsection{The Proposed Method.}
To generate viewpoint-transformed samples, 
we draw on the idea of spatio-temporal rotation.
For viewpoint transformation~(\textbf{VPT}), 
we introduce translation matrices $T_b$, $T_a$, 
which represent the translation to the rotation center
$(x_c, y_c, t_c)$
and the translation 
back to the original position, respectively.
\begin{equation}\label{eq_matrix_T}
  \begin{aligned}
    T_{b}=
    \begin{pmatrix}
      1 & 0 & 0 & 0\\
      0 & 1 & 0 & 0\\
      0 & 0 & 1 & 0\\
      -y_c & -x_c & -t_c & 1 \\
    \end{pmatrix}
  \end{aligned},
  \begin{aligned}
    T_{a}=
    \begin{pmatrix}
      1 & 0 & 0 & 0\\
      0 & 1 & 0 & 0\\
      0 & 0 & 1 & 0\\
      y_c & x_c & t_c & 1 \\
    \end{pmatrix}
  \end{aligned}
\end{equation}

Suppose that 
rotate along the $y$ and $t$ planes with $x$ as the axis,
we can easily derive the rotation matrix $R_{r}^{YT}$:
\begin{equation}
\begin{aligned}
  R_{r}^{YT}=
  \begin{pmatrix} 
    cos\theta & 0 & sin\theta &0\\
    0 & 1 & 0 &  0\\
    -sin\theta & 0 & cos\theta & 0\\
    0 & 0 & 0 & 1 \\
  \end{pmatrix}
\end{aligned}\label{eq_unbalancedYT}
\end{equation}
where $\theta$ is the rotation angle.
In practice, Eq~\ref{eq_unbalancedYT} is an 
unbalanced matrix due to the mismatch between the time and space dimensions 
in the $M_{\varepsilon}$ matrix.
Therefore, we introduce a balance coefficient $\tau$ to scale 
the space and time dimension, which results in
a better visual effects.
The balanced matrix $R_{br}^{YT}$ 
can be formulated as:
\begin{equation}
  \begin{aligned}
    R_{br}^{YT}=
    \begin{pmatrix} 
      cos\theta & 0 &\tau sin\theta &0\\
      0 & 1 & 0 &  0\\
      -\frac{1}{\tau}sin\theta & 0 & cos\theta & 0\\
      0 & 0 & 0 & 1 \\
    \end{pmatrix}
  \end{aligned}\label{eq_balancedYT}
\end{equation}

\begin{table*}[t] 
  \centering
  \caption{Performance of VPT-STS on SNNs and CNNs with various representations.}\label{tab_SNN_ANN} 
  \begin{tabular}{ccccccc}
  \toprule
  \multirow{2}{*}{Datasets}  & \multirow{2}{*}{Method} & \multicolumn{5}{c}{Accuracy (\%)} \\ \cmidrule(l){3-7} 
                              &                         & SNNs   & EventFrame & EventCount  & VoxelGrid   & EST \\ \midrule
  \multirow{2}{*}{CIFAR10-DVS}   & Baseline              & 83.20 & 78.71& 78.85 & 77.47 & 78.81 \\ 
                                 & VPT-STS         & 84.40 & 79.58& 79.12 & 79.62 & 79.37 \\ \midrule
  \multirow{2}{*}{N-Caltech101}  & Baseline              & 78.98 & 73.08& 73.66 & 77.08 & 78.41 \\ 
                                 & VPT-STS        & 81.05 & 76.96& 76.38 & 79.13 & 78.88 \\ \midrule
  \multirow{2}{*}{N-CARS}        & Baseline              & 95.40 & 94.44& 94.76 & 93.86 & 94.97 \\ 
                                 & VPT-STS        & 95.85 & 94.60& 94.81 & 94.30 & 94.99 \\ \bottomrule
  \end{tabular}\vspace*{-10pt}
\end{table*}

Set $x_c = 0$, the viewpoint transformation matrix 
$M_{br}^{YT}$ can be formulated 
by calculating $T_bR_{br}^{YT}T_a$:
\begin{footnotesize}
  \begin{equation}\label{eq_matrix_MYT}
    \begin{aligned}
      \begin{pmatrix}
        cos\theta & 0 & \tau sin\theta & 0\\
        0 & 1 & 0 & 0\\
        -\frac{1}{\tau}sin\theta & 0 & cos\theta & 0\\
        -x_ccos\theta+\frac{1}{\tau}t_csin\theta+x_c & 0 & -\tau x_csin\theta-t_ccos\theta+t_c & 1 \\
      \end{pmatrix}
    \end{aligned}
  \end{equation}
\end{footnotesize}

Similarly, the viewpoint transformation matrix $M_{br}^{XT}$ 
in the $x$ and $t$ dimensions can be formulated as:
\begin{footnotesize}
  \begin{equation}\label{eq_matrix_MXT}
    \begin{aligned}
      \begin{pmatrix}
        1 &0 & 0 & 0\\
        0 & cos\theta & \tau sin\theta & 0\\
        0 & -\frac{1}{\tau}sin\theta & cos\theta & 0\\
        0 & -x_ccos\theta+\frac{1}{\tau}t_csin\theta+x_c & -\tau x_csin\theta-t_ccos\theta+t_c & 1 \\
      \end{pmatrix}
    \end{aligned}
  \end{equation}
\end{footnotesize}

Therefore, the viewpoint-transformed matrix 
$M_{VPT}^{YT}$ and $M_{VPT}^{XT}$ 
can be formulated as:
\begin{equation}\label{eq_matrxVPT}
  \left.\begin{aligned}
    M_{VPT}^{YT} = M_{\varepsilon}M_{br}^{YT}&\\
    M_{VPT}^{XT} = M_{\varepsilon}M_{br}^{XT}& 
  \end{aligned} \right\} 
\end{equation}

Furthermore, since events beyond the resolution will 
be discarded during the viewpoint transformation, 
we introduce spatiotemporal stretching~(\textbf{STS}) to 
avoid potential information loss. 
STS stretches the temporal mapping in the VPT 
by a coefficient $\frac{1}{cos\theta}$ 
while maintaining the spatial coordinates unchanged.
Therefore, by setting $t_c= 0$, we get the transformed 
$(t)_{STS}^{YT}$ and $(t)_{STS}^{XT}$ 
from  Eq.~\ref{eq_matrix_MYT} and
Eq.~\ref{eq_matrix_MXT}:
\begin{equation}\label{eq_t_STS}
  \begin{aligned}
    \left.\begin{aligned}
      (t_k)_{VPT}^{YT} = (t_k)-\tau tan\theta \cdot ((y_k)-y_c)&\\
      (t_k)_{VPT}^{XT} = (t_k)-\tau tan\theta \cdot ((x_k)-x_c)& 
    \end{aligned} \right\} 
  \end{aligned}
\end{equation}

The time of STS is advanced 
or delayed according to the distance from 
the center $|x-x_c|$~($|y-y_c|$),
causing event stream to be stretched 
long the time axis according to the spatial coordinates.

\section{Experiments}\label{sec:experiments}
\begin{figure*}[tb]
  \centering
  \subfigure[Baseline accuracy.]{\includegraphics[width=0.3\linewidth]{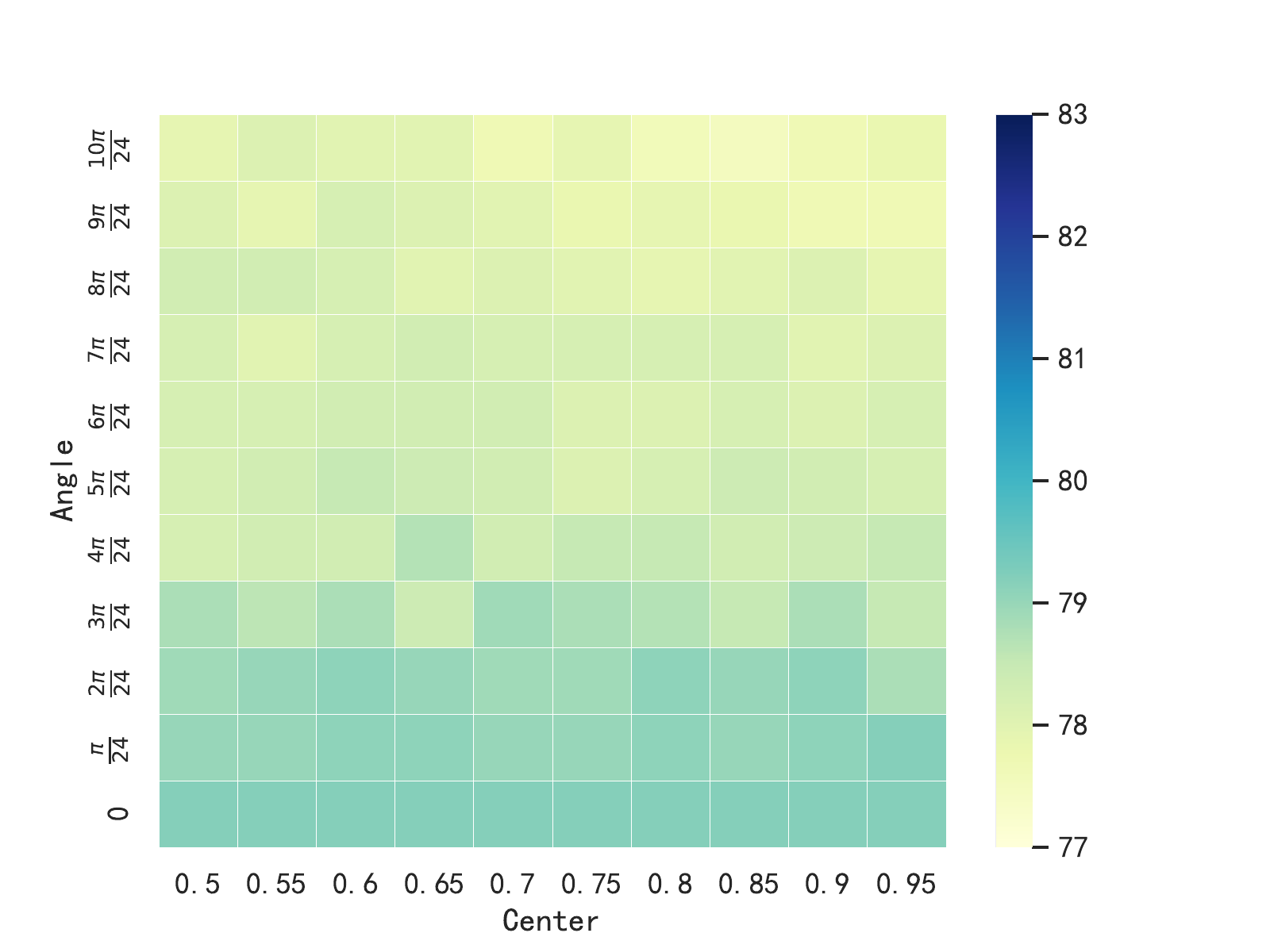}\label{fig_baseline}}  
  \subfigure[VPT-STS accuracy.]{\includegraphics[width=0.3\linewidth]{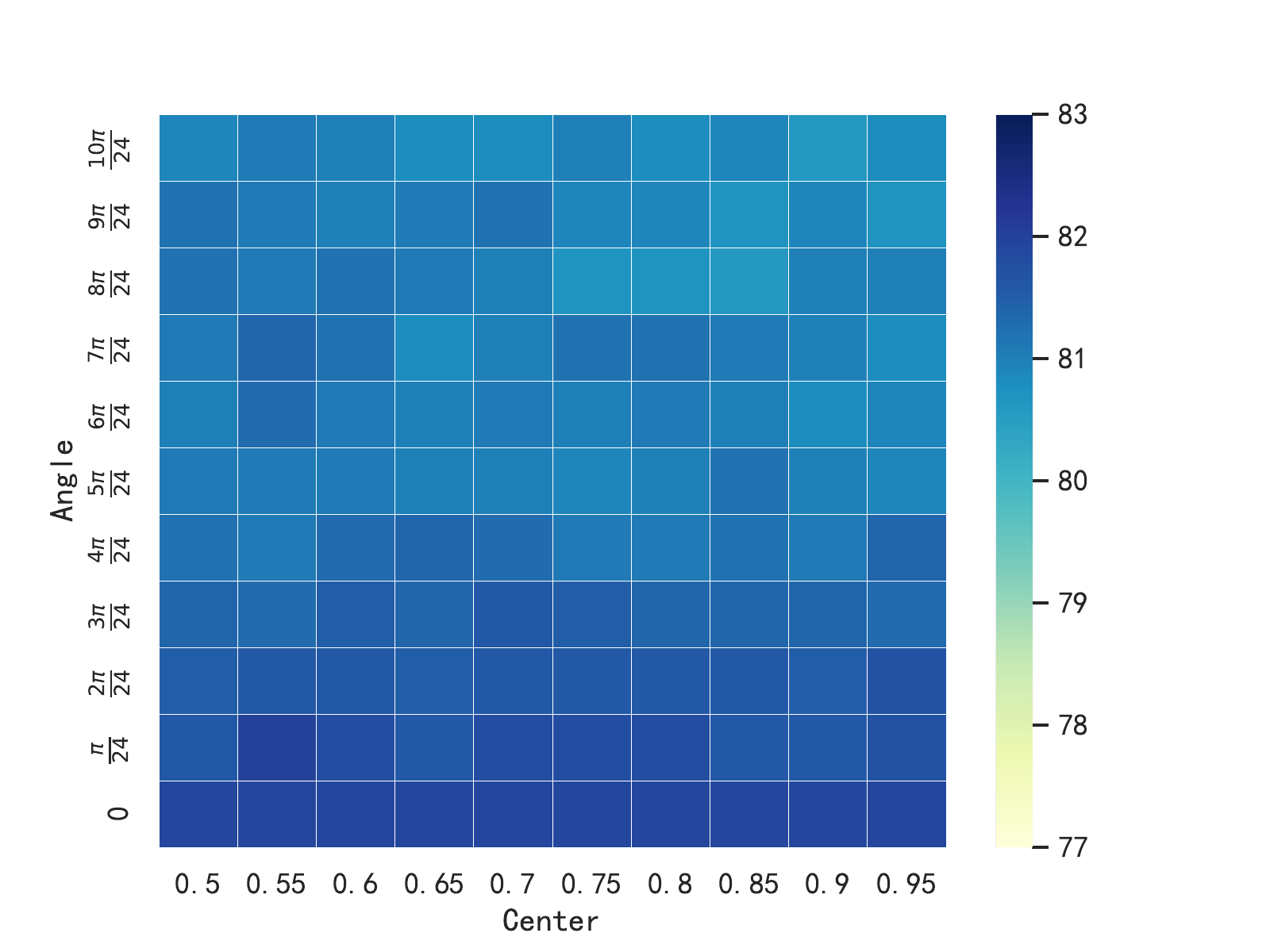}\label{fig_eventr}}
  \subfigure[Difference of accuracy reduction.]{\includegraphics[width=0.3\linewidth]{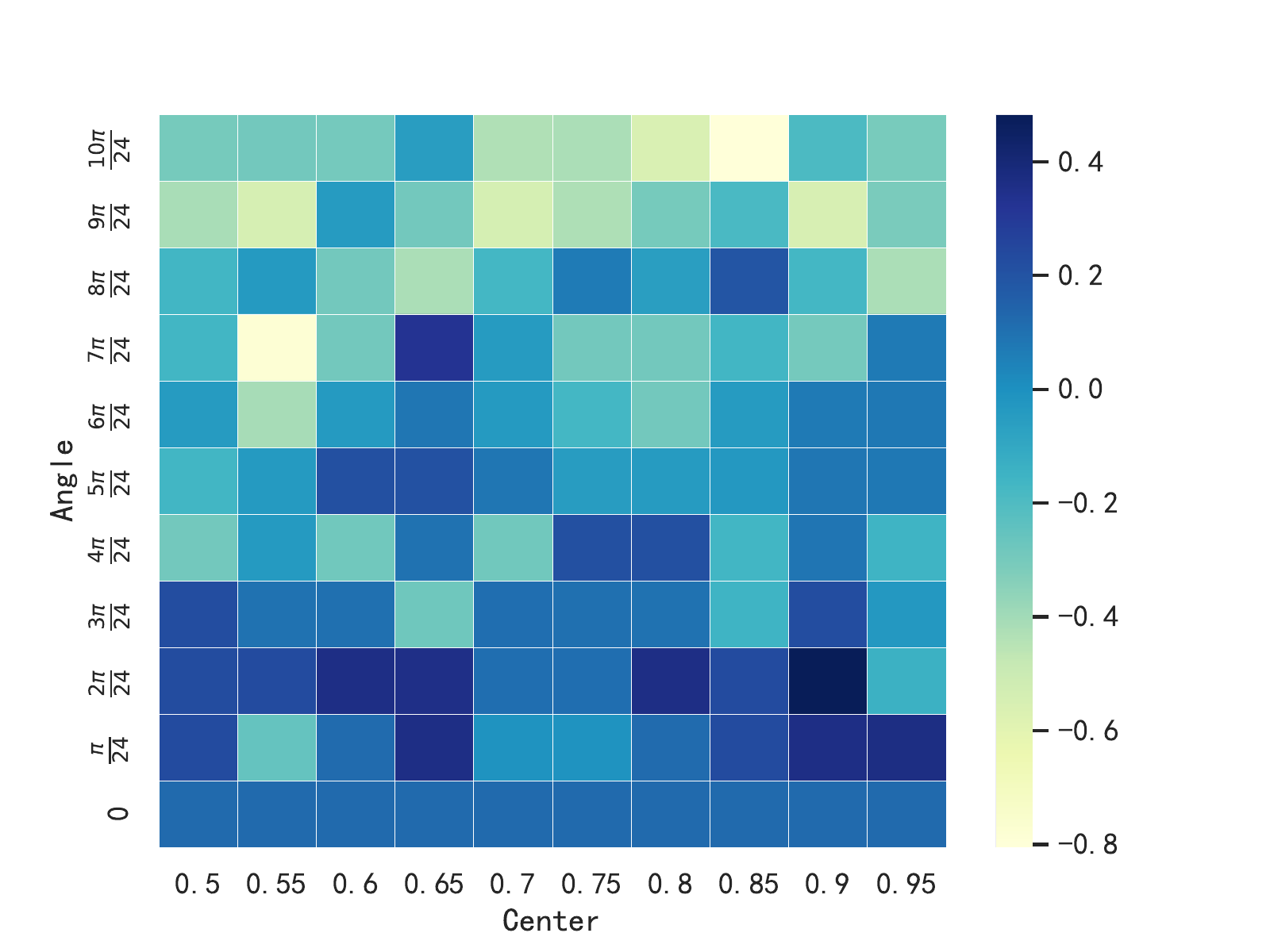}\label{fig_difference}}
  \caption{Performance of VPT-STS and Baseline under different perturbations.}
  \label{fig_analysis} 
  \vspace*{-10pt}
\end{figure*}
% In this section, we verify the effectiveness and applicability of our approach. In Section 
% 4.1, we introduce the dataset we adopted as well as the experimental configuration. In Section 
% 4.2, we present a comparison with the latest state-of-the-art literature. In Section 4.3, 
% we present the analysis in full range on different networks. In Section 4.3, we evaluate 
% the effectiveness of different strategies.
\subsection{Implementation}
Extensive experiments are performed to demonstrate 
the superiority of the VPT-STS method on prevailing 
neuromorphic datasets, including CIFAR10-DVS(CIF-DVS)~\cite{DVSCIFAR10}, 
N-Caltech101(N-Cal)~\cite{NCALTECH101NMNIST}, N-CARS~\cite{NCARS} datasets. 
N-Caltech101 and CIFAR10-DVS datasets are generated 
by neuromorphic vision sensors on the basis of traditional datasets, 
while N-CARS is collected in the real world. 
For the convenience of comparison, 
the model without VPT-STS with the same parameters 
is used as the baseline. 
STBP~\cite{zheng2021going} methods 
are used to train SNN-VGG9 network, 
other parameters mainly refer to NDA~\cite{li2022neuromorphic}. 
For example, the Adam optimizer is used with 
an initial learning rate of $1e-3$. 
The neuron threshold and leakage coefficient are $1$ and $0.5$, 
respectively. 
In addition, we also evaluate the performance of 
VPT-STS on various event representations 
with the Resnet9 network, including 
EST~\cite{EST}, VoxelGrid~\cite{VoxelGrid}, 
EventFrame~\cite{EventFrame} and 
EventCount~\cite{EventCount} representations.

\begin{table}[tb]
  \centering
  \caption{Performance of VPT-STS and previous SOTAs on CIFAR10-DVS and N-CARS datasets.}\label{tab_sota_acc}
  \begin{tabular}{cccc}
      \toprule 
      \multirow{2}{*}{Methods} &\multirow{2}{*}{References}&\multicolumn{2}{c}{Accuracy (\%)} \\ \cmidrule(l){3-4}
                    &                   &CIF-DVS    & N-CARS\\  
      \midrule
      HATS\cite{NCARS}                  &CVPR 2018 & 52.40 & 81.0 \\
      Dart\cite{RameshYOTZX2020}        &TPAMI 2020 & 65.80 &  -  \\ 
      Dspike~\cite{li2021differentiable}&NeurIPS 2021 &75.40 & -\\ 
      STBP~\cite{zheng2021going}        &AAAI 2021 &67.80 & -\\ 
      AutoSNN~\cite{na2022autosnn}      &ICML 2022 &72.50& -\\
      RecDis~\cite{guo2022recdis}       &CVPR 2022 &72.42& -\\
      DSR~\cite{meng2022training}       &CVPR 2022 &77.27& -\\ 
      NDA~\cite{NDA}                    &ECCV 2022 &81.70 &90.1 \\ 
      \midrule
      \textbf{VPT-STS}    &- &\textbf{84.40} &\textbf{95.85}\\ 
      \bottomrule
  \end{tabular}
\end{table}

\subsection{Performance on various representations}
Extensive experiments are conducted to evaluate the performance of 
VPT-STS method on different event representations, 
covering SNNs and CNNs. 
As shown in Tab.~\ref{tab_SNN_ANN}, 
SNNs with VPT-STS methods achieve significant 
improvements on three prevailing datasets. 
And VPT-STS also performs well on four 
representations commonly used by CNNs. 
It is worth noting that EST maintains the most spatiotemporal information 
from neuromorphic data and thus performs best overall. 
Furthermore, since the samples of N-CARS are collected 
in the real world, 
its initial viewpoint diversity is already enriched 
compared to the other two datasets. 
Considering the high baseline on N-CARS, 
VPT-STS still further imporves the robustness of SNNs.

\begin{table}[b]
  \vspace*{-10pt}
  \centering
  \caption{Comparison of Different Strategies.}\label{tab_Strategies}
  \begin{tabular}{cccc}
      \toprule 
      % acc, CNN, SNN
      \multirow{2}{*}{Methods} & \multicolumn{3}{c}{Accuracy (\%)} \\ \cmidrule(l){2-4}
                          & CIF-DVS   & N-Cal  & N-CARS\\ \midrule 
      Baseline            & 83.20         &  78.98        & 95.40  \\ 
      Rotation            & 83.90         &  80.19        & 95.46  \\ 
      VPT           & \textbf{84.40}& \textbf{81.05}& 95.56  \\ 
      STS           & 84.30         &  80.56        & \textbf{95.85} \\ \bottomrule
  \end{tabular}
  \vspace{-10pt}
\end{table}
\subsection{Compared with SOTAs}
As shown in Tab.~\ref{tab_sota_acc}, 
we compare VPT-STS with recent state-of-the-art results 
on neuromorphic datasets. 
The results show that VPT-STS achieves substantial 
improvements over previous SOTAs. 
It is worth noting that VPT-STS significantly outperforms NDA, 
which is an ensemble of six geometric transformations. 
The experimental results demonstrate the superiority of 
combining spatiotemporal information for data augmentation.
Since VPT-STS is orthogonal to most training algorithms, 
it can provide a better baseline and improve 
the performance of existing models.

\subsection{Ablation Studies on VPT-STS}
\begin{figure}[tb]
  \centering
    \centerline{\includegraphics[width=6.0cm]{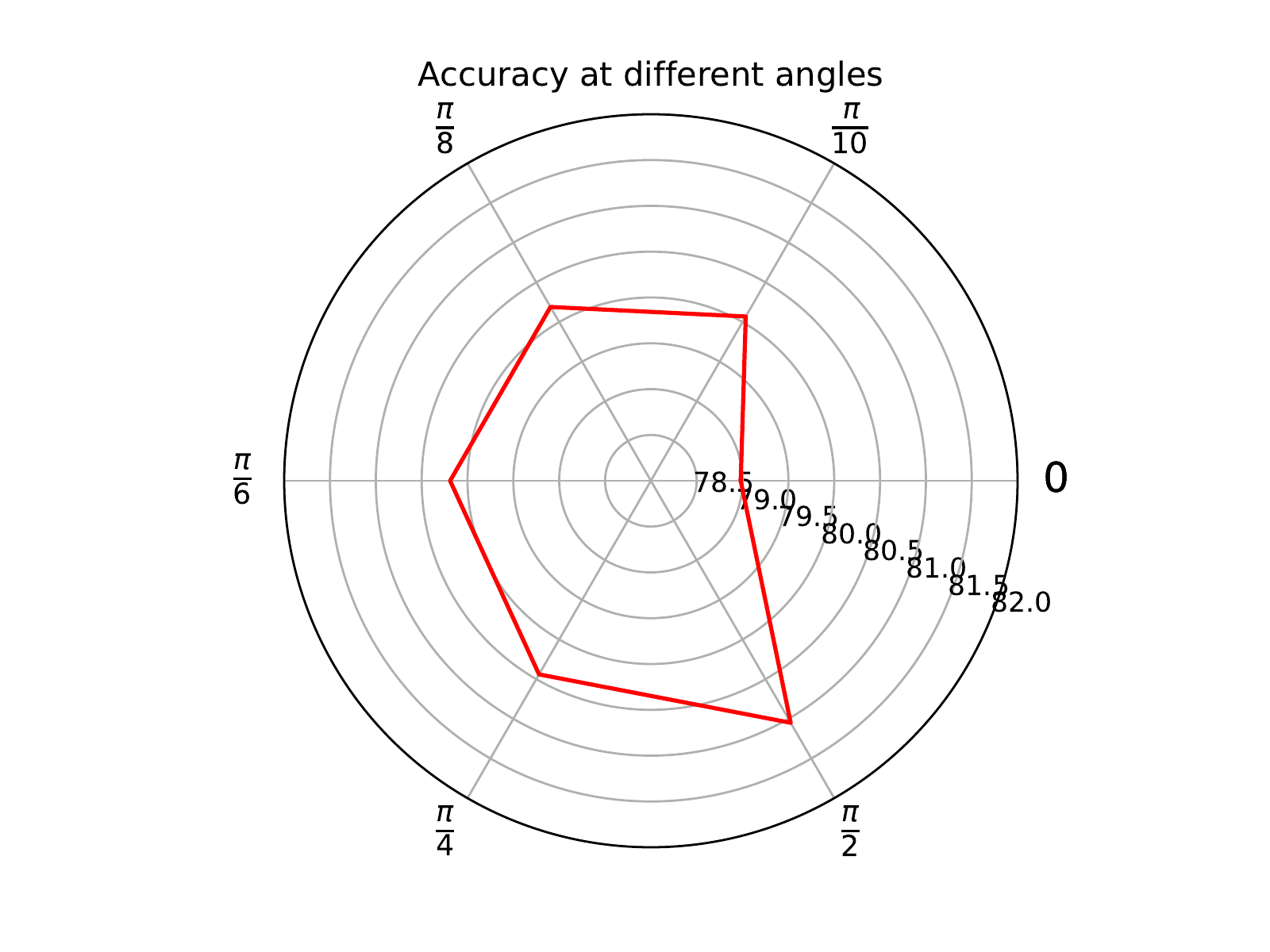}}
  \caption{Performance of VPT-STS at different angles.}
  \label{fig_Angle}
  \vspace*{-10pt}
\end{figure}

As shown in Fig.~\ref{fig_Angle}, the performance of 
VPT-STS with different rotation angles is evaluated 
on the N-Caltech101 dataset. 
It turns out that a suitable rotation angle is 
important for the performance of data augmentation, 
which can increase data diversity without losing features.

\subsection{Analysis of VPT-STS}
To gain further insight into the workings of VPT-STS,
we add different strategies on the baseline to analyze 
the effective components of VPT-STS.
As shown in Table~\ref{tab_Strategies}, 
spatial rotation (Rotation) is performed as a comparative experiment for VPT-STS. 
It turns out that both VPT and STS including spatiotemporal transformations 
are significantly better than pure spatial geometric transformations 
on all three datasets, 
which illustrate the importance of spatiotemporal transformations. 
While VPT and STS are implemented with operations similar to rotation, 
it actually improves the robustness of SNNs to different viewpoints.
Furthermore,
we evaluate the robustness of SNNs to viewpoint fluctuations
by adding different degrees of spatiotemporal rotation to the test data. 
% The angular range of the perturbation is $[0, \frac{5}{6}\pi]$, 
% and the range of the rotation center is $[0.5, 0.95]$. 
Figures~\ref{fig_baseline} and~\ref{fig_eventr} show 
the performance of the baseline model and 
the model trained by VPT-STS under different disturbances, respectively. 
The results show that the general trend of the accuracy change 
is to decrease with the increase of the perturbation amplitude. 
In addition, Fig.~\ref{fig_difference} shows the difference 
in the accuracy reduction of the VPT-STS compared to baseline. 
As the perturbation amplitude increases, 
the difference in the accuracy reduction of the two models 
is less than zero, and the absolute value grows, 
which illustrate that the accuracy reduction of baseline is larger 
than that of VPT-STS. 
Experimental results show that the model trained 
with VPT-STS generalize better and improves 
the robustness of SNNs against spatial location variances.

\vspace*{-10pt}
\section{Conclusion}\label{sec:conclusion}
We propose a novel data augmentation method 
suitable for events, 
viewpoint transformation and spatiotemporal stretching~(VPT-STS). 
Extensive experiments on prevailing neuromorphic datasets 
show that VPT-STS is broadly effective 
on multiple event representations and 
significantly outperforms pure spatial geometric transformations. 
It achieves substantial improvements over previous SOTAs 
by improving the robustness of SNNs to different viewpoints.

\newpage
% References should be produced using the bibtex program from suitable
% BiBTeX files (here: strings, refs, manuals). The IEEEbib.bst bibliography
% style file from IEEE produces unsorted bibliography list.
% -------------------------------------------------------------------------
\bibliographystyle{IEEEbib}
\bibliography{strings}

\end{document}